\documentclass[english,10pt,twocolumn,letterpaper]{article}
\usepackage[latin9]{inputenc}
\usepackage{array}
\usepackage{amsmath}
\usepackage{amssymb}
\usepackage{graphicx}
\usepackage{setspace}

\makeatletter

\providecommand{\tabularnewline}{\\}

 \usepackage{cvpr2}
 \usepackage{times}
 \usepackage{amsmath}
 \usepackage{amssymb}
\def\cvprPaperID{1888}
\cvprfinalcopy
\usepackage{etoolbox} 
\patchcmd\thebibliography  {\labelsep}  {\labelsep\itemsep=0pt\relax}  {}  {\typeout{Couldn't patch the command}}

\makeatother

\usepackage{babel}
\begin{document}

\title{Object Detection with Mask-based Feature Encoding}

\author{%
\begin{tabular}{>{\centering}p{3cm}||>{\centering}p{3cm}||>{\centering}p{3cm}||>{\centering}p{3cm}}
\multicolumn{4}{c}{Xiaochuan Fan$^{1}$,$\ $ Hao Guo$^{2}$,$\ $ Kang Zheng$^{2}$,
$\ $ Wei Feng$^{3}$, $\ $Song Wang$^{2}$}\tabularnewline
\multicolumn{4}{>{\centering}p{14cm}}{$^{1}$ HERE North America LLC}\tabularnewline
\multicolumn{4}{c}{$^{2}$ Department of Computer Science \& Engineering, University
of South Carolina}\tabularnewline
\multicolumn{4}{c}{$^{3}$ School of Computer Science and Technology, Tianjin University}\tabularnewline
\multicolumn{4}{c}{\tt{\small{}efan3000@gmail.com, \{hguo,zheng37\}@email.sc.edu, wfeng@ieee.org, songwang@cec.sc.edu}}\tabularnewline
\end{tabular}}

\maketitle
\thispagestyle{empty}
\begin{abstract}
Region-based Convolutional Neural Networks (R-CNNs) have achieved
great success in the field of object detection. The existing R-CNNs
usually divide a Region-of-Interest (ROI) into grids, and then localize
objects by utilizing the spatial information reflected by the relative
position of each grid in the ROI. In this paper, we propose a novel
feature-encoding approach, where spatial information is represented
through the spatial distributions of visual patterns. In particular,
we design a Mask Weight Network (MWN) to learn a set of masks and
then apply channel-wise masking operations to ROI feature map, followed
by a global pooling and a cheap fully-connected layer. We integrate
the newly designed feature encoder into the Faster R-CNN architecture.
The resulting new Faster R-CNNs can preserve the object-detection
accuracy of the standard Faster R-CNNs by using substantially fewer
parameters. Compared to R-FCNs using state-of-art PS ROI pooling and
deformable PS ROI pooling, the new Faster R-CNNs can produce higher
object-detection accuracy with good run-time efficiency. We also show
that a specifically designed and learned MWN can capture global contextual
information and further improve the object-detection accuracy. Validation
experiments are conducted on both PASCAL VOC and MS COCO datasets.
\end{abstract}

\section{Introduction}

\begin{figure}
\begin{centering}
\includegraphics[scale=0.33]{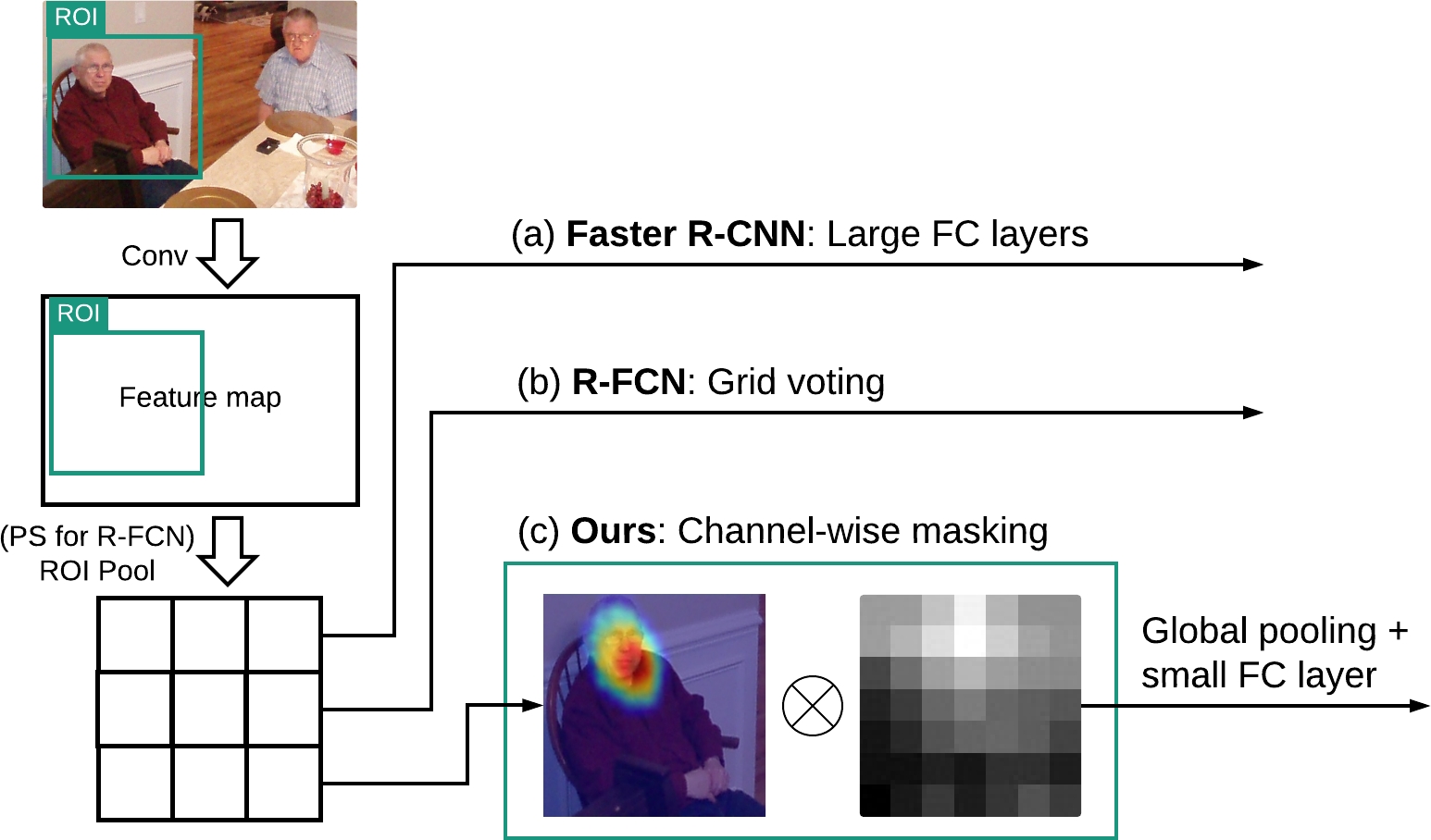}
\par\end{centering}

\caption{Feature-encoding comparison between our approach and Faster R-CNN/R-FCN.
(a) Faster R-CNN: ROI pooling followed by large FC layers. (c) R-FCN:
PS ROI pooling followed by voting. (c) Ours: ROI pooling followed
by channel-wise masking, global pooling and a small FC layer. \label{fig:doorbuster}}
\end{figure}

Region-based convolutional neural networks (R-CNNs) have been recognized
as one of the most effective tools for object detection \cite{Girshick2014,Girshick2015,Ren2015,Lin2016,Dai2016,Dai2017,He2017}.
One important component in R-CNNs is region-wise feature encoder.
In the standard Fast/Faster R-CNN \cite{Girshick2015,Ren2015} architectures,
this component is implemented by the Region-of-Interest (ROI) pooling
followed by fully-connected (FC) layers, as shown in Fig. \ref{fig:doorbuster}(a).
Such a feature encoder usually introduces a huge number of connections
and parameters. Recently, very deep networks, such as ResNet (Residual
Network)  \cite{He2016} and SENet (Squeeze-and-Excitation Network)
\cite{Hu2017}, have been proposed for image classification, without
the inclusion of large FC layers. However, it is non-trivial to use
these networks to help improve the object-detection performance due
to the lack of translation variance \cite{Dai2016}. Special architectures
have to be used in feature encoder to encode spatial information into
these networks for object detection \cite{Dai2016,Dai2017}, e.g.
PS (Position-Sensitive) ROI pooling in R-FCN (Region-based Fully Convolutional
Network) as shown in Fig. \ref{fig:doorbuster}(b). To our best knowledge,
all these architectures utilize grids to represent object-parts and
reflect the spatial information in the ROI. 

In this paper, a novel feature-encoding approach is presented for
object detection. The basic idea behind our approach is that, compared
with grids, it is more natural to use middle-level visual patterns
to represent object-parts. Given that each channel of a CNN feature
map is expected to be an activation map for a specific visual pattern,
e.g., middle-level attribute or object part \cite{Zeiler2014,Zhou2016},
we propose to learn a set of masks to reflect the spatial distribution
associated to these visual patterns, and then applying channel-wise
masking operations to ROI feature map, followed by a global pooling
and a cheap fully-connected layer. This architecture is illustrated
in Fig. \ref{fig:doorbuster}(c), where, as an example, a feature
channel which is strongly activated for human head is masked by a
learned mask. We design a Mask Weight Network (MWN) for mask learning,
and MWN is jointly trained with the whole object-detection network.

To validate the effectiveness and efficiency of proposed new feature
encoder, we integrate it into the Faster R-CNN architecture and find
that the resulting new Faster R-CNNs, named \textit{MWN-based Faster
R-CNN} (M-FRCN) in this paper, perform very well in terms of object-detection
accuracy, model complexity, and run-time efficiency. More specifically,
as compared with the standard Faster R-CNN with large FC layers, our
M-FRCN preserves the object-detection accuracy, but using substantially
fewer parameters. As compared with R-FCNs using state-of-art PS ROI
pooling \cite{Dai2016} and deformable PS ROI pooling \cite{Dai2017},
our M-FRCN can produce higher object-detection accuracy without losing
time efficiency. We also show that a specifically designed and learned
MWN can capture global contextual information. The combination of
two MWNs, one for local regions and the other for global context,
leads to new Faster R-CNNs, that can further improve the object detection
accuracy. We conduct the validation experiments on both PASCAL VOC
and MS COCO datasets. 

\begin{figure*}
\begin{centering}
\includegraphics[scale=0.63]{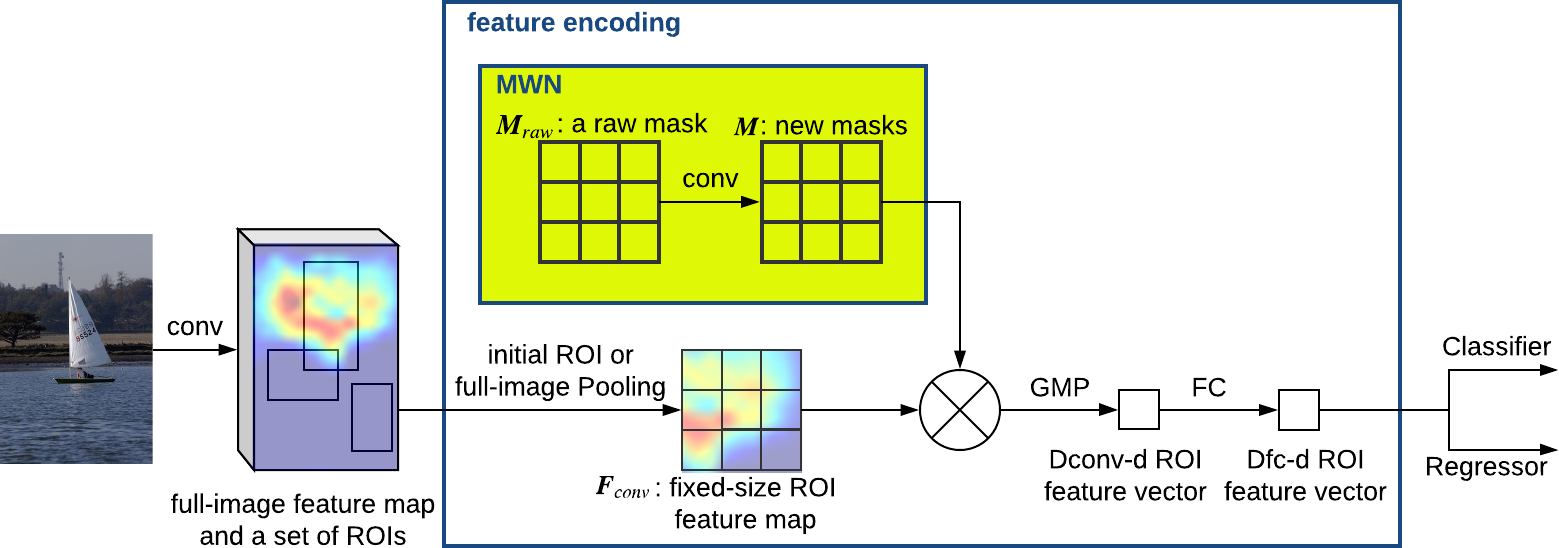}
\par\end{centering}

\caption{Architecture of the proposed MWN-based Faster R-CNN (M-FRCN). \label{fig:M-FRCN}}
\end{figure*}

\section{Related Work}

Girshick et al. \cite{Girshick2014} first propose R-CNNs by evaluating
CNNs on region proposals for bounding-box object detection. \cite{Girshick2015}
extends R-CNN to Fast R-CNN by applying ROI pooling to enable end-to-end
detector training on shared convolutional features. Ren et al. \cite{Ren2015}
further extend Fast R-CNN to Faster R-CNN by incorporating a Region
Proposal Network (RPN). In the standard Fast/Faster R-CNN, the ROI
pooling layer is followed by large fully-connected layers.

Recently, very deep feature-extraction architectures are proposed
\cite{Lin2014,He2016,Szegedy2015,Iandola2016,Howard2017,Huang2017,Hu2017},
which do not need large fully-connected layers any more, such as GoogLeNet,
ResNet, SENet, and DenseNet. Although these very-deep architectures
can achieve impressive image-classification accuracy, it is non-trivial
to directly use them to improve object detection due to their lack
of spatial information encoding, as introduced in \cite{Dai2016}.
To address this issue, He et al. \cite{He2016} insert the ROI pooling
layer into convolutions to introduce translation variance. Dai et
al. \cite{Dai2016} propose R-FCN by adding a position-sensitive ROI
pooling. In DCN (Deformable Convolutional Networks) \cite{Dai2017},
a deformable ROI pooling is proposed to add a learned offset to each
ROI grid position, and thus enables adaptive part localization. In
this paper, we propose a new feature-encoding approach for object
detection. In the experiments, we show that this new feature encoder
can be used for both shallow networks, e.g. VGG\_CNN\_M\_1024 \cite{Chatfield2014},
and very deep networks, e.g. ResNet-101 \cite{He2016}, to improve
object detection in terms of detection accuracy, number of parameters,
and/or run-time efficiency.

Masks have been widely used in a variety of visual tasks, such as
object detection \cite{Gidaris2015,Kantorov2016,He2017}, semantic
segmentation \cite{He2017,Dai2015,Pinheiro2015}, and pose estimation
\cite{He2017,Fan2015a}. Gidaris and Komodakis \cite{Gidaris2015}
and Kantorov et al. \cite{Kantorov2016} use a set of manually-designed
masks to represent specific spatial patterns. \cite{Pinheiro2015}
and \cite{He2017} propose models to estimate segmentation masks.
In \cite{Dai2015}, binary masks are applied to the convolutional
feature maps to mask out background region. In this paper we learn
masks to capture spatial information for object detection.

Global contextual information has been proven to be very valuable
for object detection and recognition \cite{Torralba2003,Rabinovich2007,Oliva2007,Mottaghi2014}.
Several techniques have been developed to incorporate global context
into CNNs. For example, ParsetNet \cite{Liu2016} concatenates features
from the full image to each element in a feature map. ResNet \cite{He2016}
performs a global ROI pooling to obtain global features and concatenates
global features to local region features. ION \cite{Bell2016} uses
a stacked spatial RNN to exploit global context. In ParsetNet and
ResNet, all local features share the same global feature. Different
from these networks, in this paper, we propose a method to extract
ROI-specific global features. Unlike ION, the proposed method uses
a convolution-based solution.

\section{Our Approach\label{sec:Our-Approach}}

In this section, we first introduce the general architecture of the
proposed MWN(Mask Weight Network)-based Faster R-CNN (M-FRCN). Then,
we describe the new MWN to learn masks for local ROIs and global context,
leading to MWN-l and MWN-g, respectively. For convenience, the proposed
M-FRCN using MWN-l and MWN-g are abbreviated as M-FRCN-l and M-FRCN-g,
respectively. Finally, we combine M-FRCN-l and M-FRCN-g into M-FRCN-lg
by integrating local region and global contextual information for
object detection. While in this paper we incorporate the proposed
feature encoders into the Faster R-CNN, it is easy to incorporate
them into other R-CNN frameworks.

\subsection{General Architecture of M-FRCN\label{sub:M-FRCN}}

As illustrated by Fig. \ref{fig:M-FRCN}, for each ROI proposed by
a Region Proposal Network (RPN) \cite{Ren2015}, we take the following
steps to encode its feature map for classification and bounding-box
regression. First, an initial $N'\times N'$ ROI (or image) pooling
converts an ROI (or image) to a fixed-size feature map $\boldsymbol{F}_{conv}\in\mathbb{R}^{N'\times N'\times D_{conv}}$,
where, $D_{conv}$ is the number of channels for $\boldsymbol{F}_{conv}$,
e.g. $D_{conv}=2048$ for a ResNet-101 model \cite{He2016}. In the
following sections, we will elaborate on the selection of the initial-pooling
input, which can be either the full image or an ROI on the image.
Different from standard Faster R-CNN, we use an average ROI pooling
instead of a max one. 

Second, a Mask Weight Network (MWN) takes a raw mask $\boldsymbol{M}_{raw}\in\mathbb{R}^{N'\times N'}$
as input and performs a set of convolution operations on this raw
mask to get a set of new masks $\boldsymbol{M}^{k}\in\mathbb{R}^{N'\times N'}$,
$k=1,2,\cdots,D_{conv}$, using $D_{conv}$ kernels and a stride equal
to $1$. We will elaborate on the selection of the raw mask and the
MWN in the following two sections.

Third, we apply the $k$-th new mask $\boldsymbol{M}^{k}$ to $\boldsymbol{F}_{conv}^{k}\in\mathbb{R}^{N'\times N'}$,
i.e., the $k$-th channel of $\boldsymbol{F}_{conv}$, by 
\begin{equation}
\begin{split}\boldsymbol{F'}_{conv}^{k}(i,j) & =\boldsymbol{F}_{conv}^{k}(i,j)\times\boldsymbol{M}^{k}(i,j),\\
i,j & =1,\ldots,N',\, k=1,\ldots,D_{conv},
\end{split}
\label{eq:masking}
\end{equation}
 where $i$ and $j$ are the horizontal and vertical positions in
a mask or feature map, and $\boldsymbol{F'}_{conv}$ denotes the ROI
feature map after masking.

Finally, a global max pooling (GMP) is performed on each channel of
$\boldsymbol{F'}_{conv}$, followed by a fully-connected (FC) layer.
We denote the number of output nodes of this FC layer as $D_{fc}$,
and thus the number of connections of the FC layer is $D_{conv}\times D_{fc}$.
Note that this new architecture does not incur large FC layers as
the standard Faster R-CNN, where $N\times N$ ROI pooling is followed
by two FC layers. Taking an example of $D_{fc}=256$ for VGG-16 architecture,
the FC layer of the proposed M-FRCN only has $512\times256=131,072$
connections while the standard Faster R-CNN has $7\times7\times512\times4096=102,760,448$
connections when setting $N=7$.

\subsection{M-FRCN-l \label{sub:M-FRCN-l}}

\begin{figure}
\begin{centering}
\includegraphics[scale=0.51]{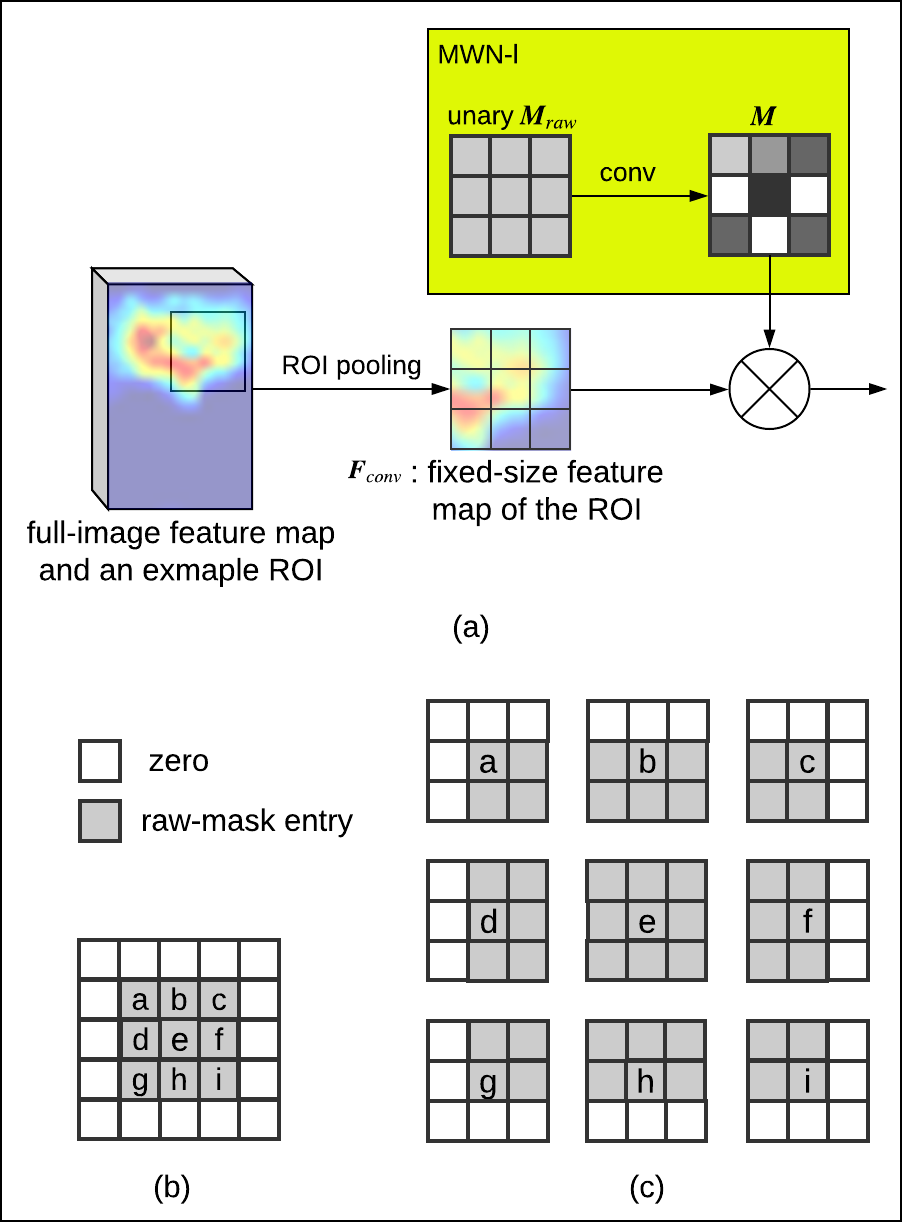}
\par\end{centering}

\caption{(a) An illustration of masking operation in M-FRCN-l with $N'=3$.
(b) The zero-padded $3\times3$ raw mask for MWN-l convolution. (c)
Each mask entry shows different pattern when considering spatial neighbors
for ($3\times3$) convolution filtering. \label{fig:mwn-l}}
\end{figure}

Following the general architecture of M-FRCN in Fig. \ref{fig:M-FRCN},
in this section we introduce M-FRCN-l by selecting a raw mask $\boldsymbol{M}_{raw}$
and a corresponding MWN-l that can learn new masks $\boldsymbol{M}^{k}$,
$k=1,2,\ldots,D_{conv}$ for local ROIs. As shown in Fig. \ref{fig:mwn-l}(a),
for M-FRCN-l, the input of the initial $N'\times N'$ pooling is a
considered ROI proposed by RPN. As highlighted in the yellow region
of Fig. \ref{fig:mwn-l}(a), we simply select $\boldsymbol{M}_{raw}$
to be a unary mask where all its entries take a preset constant value
of $I^{l}>0$. MWN-l is comprised of a convolutional layer with $D_{conv}$
kernels, which transforms $\boldsymbol{M}_{raw}$ to the new masks
$\boldsymbol{M}^{k}$, $k=1,2,\ldots,D_{conv}$. In our approach,
we always select the convolution kernel size equal to $N'\times N'$,
and the raw mask $\boldsymbol{M}_{raw}$ is padded with zeros in the
convolution. Fig. \ref{fig:mwn-l}(b) presents an example of the zero-padded
mask, where mask size and convolution kernel size are both equal to
$3\times3$. Note that the zero-padded mask becomes a binary mask
since $I^{l}\neq0$.

A unary raw mask with preset constant entries seems to have no meaningful
information to exploit by convolution. However, convolution at each
mask entry actually involves this entry and its spatial neighbors.
With zero-padding, each mask entry shows different spatial pattern
when considering their neighbors, as shown in Fig. \ref{fig:mwn-l}(c).
These patterns reflect the relative position of each entry in ROI.
Moreover, during the training, the network loss of classification
and regression can be propagated backwards to MWN-l to learn the parameters
in its convolution layer. In general CNNs, each channel of a CNN feature
map is expected to be an activation map for a specific visual pattern
\cite{Zeiler2014,Zhou2016}, e.g. middle-level attribute or object
part. Similarly, the new masks $\boldsymbol{M}^{k}$, $k=1,2,\ldots,D_{conv}$,
output by MWN-l can reflect the spatial distribution of the visual
pattern associated to each channel of the ROI feature map, i.e. the
convolution result on each entry of the raw mask can represent the
likelihood that a visual pattern appears at the position of that entry,
according to the training set. By applying each new mask to its corresponding
feature channel, we expect that the proposed M-FRCN can encode the
spatial information necessary for accurate object detection. In the
inference stage, the learned mask is independent with the input image
and can be considered as constant, so the convolution in MWN-l can
be waived.

\begin{figure}
\begin{centering}
\includegraphics[scale=0.51]{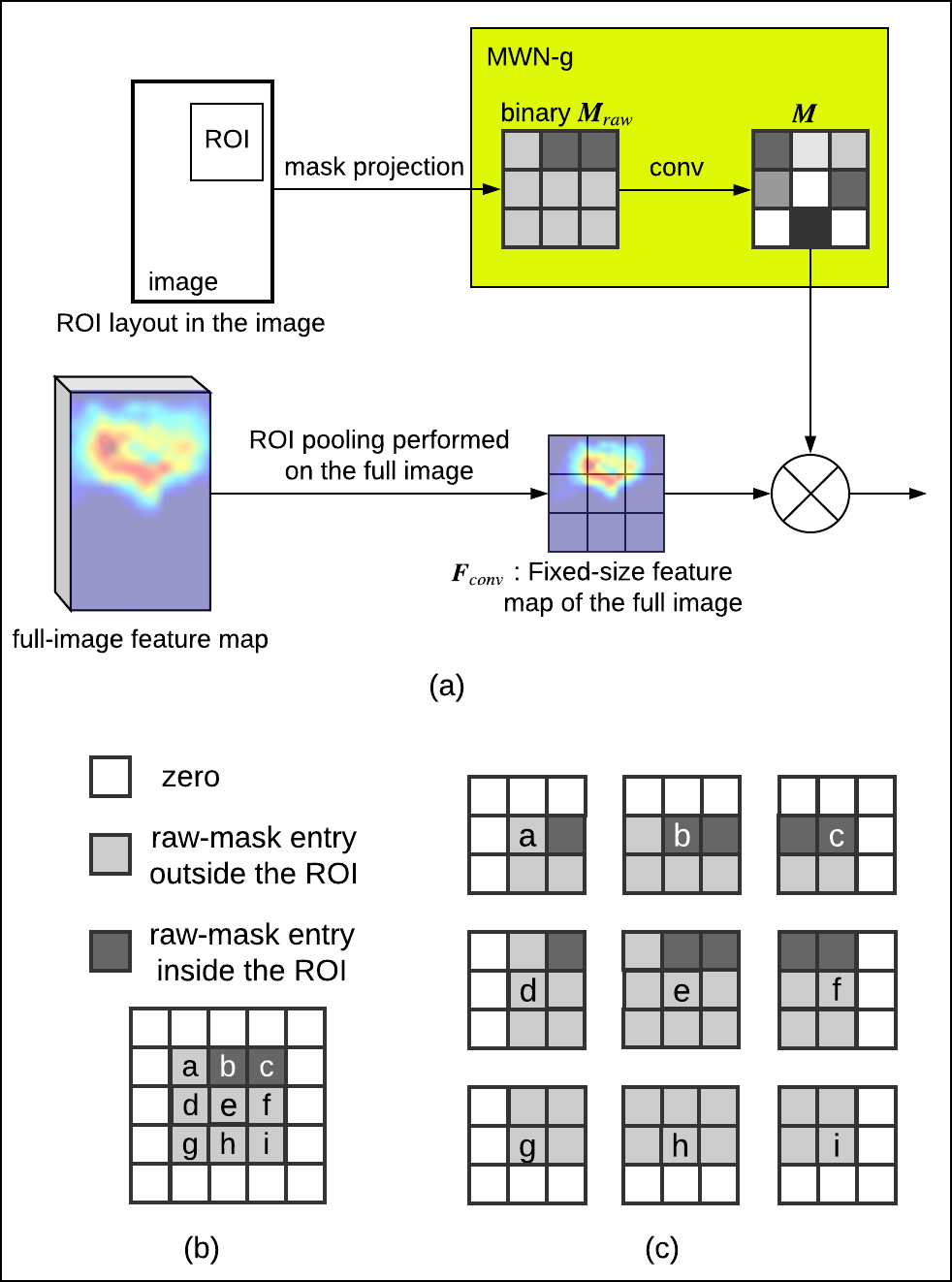}
\par\end{centering}

\caption{(a) An illustration of masking operation in M-FRCN-g with $N'=3$.
(b) A zero-padded $3\times3$ raw mask for the MWN-g convolution.
(c) Each mask entry shows different pattern when considering spatial
neighbors for ($3\times3$ ) convolution filtering. These patterns
reflect the relative position of each entry not only in the full image,
but also to the ROI. \label{fig:MWN-g}}

\centering{}
\vspace{-0.00in}
\end{figure}

\subsection{M-FRCN-g \label{sub:M-FRCN-g}}

Following the general architecture of M-FRCN in Fig. \ref{fig:M-FRCN},
in this section we introduce M-FRCN-g by selecting a raw mask $\boldsymbol{M}_{raw}$
and a corresponding MWN-g that can learn new masks $\boldsymbol{M}^{k}$,$k=1,2,\ldots,D_{conv}$
for global context. As shown in Fig. \ref{fig:MWN-g}(a), for M-FRCN-g,
the input of the initial $N'\times N'$ pooling is the full image
for the global-contextual information. We then select the raw mask
$\boldsymbol{M}_{raw}$ to reflect the relative position of a considered
ROI in the image.

More specifically, for the considered ROI, we construct a binary context
map of the same size as the image. We set an entry of this context
map to $I_{in}^{g}$ if it is located inside the considered ROI, and
$I_{out}^{g}$ otherwise, where $I_{in}^{g}\neq I_{out}^{g}$. We
then down-sample this context map to the size of $N'\times N'$ and
take it as the raw mask $\boldsymbol{M}_{raw}$ of the considered
ROI, as highlighted in the yellow region of Fig. \ref{fig:MWN-g}(a).
For MWN-g that transforms the raw mask $\boldsymbol{M}_{raw}$ to
the new masks $\boldsymbol{M}^{k}$, $k=1,2,\ldots,D_{conv}$, we
use a layer of convolution. Like MWN-l, the convolution kernel size
is always set to $N'\times N'$, and the raw mask $\boldsymbol{M}_{raw}$
is padded with zeros in the convolution. Fig. \ref{fig:MWN-g}(b)
presents an example of the zero-padded mask for a considered ROI in
MWN-g, where mask size and convolution kernel size are both equal
to $3\times3$. With zero-padding, each mask entry shows different
spatial pattern when considering their neighbors, as shown in Fig.
\ref{fig:MWN-g}(c). These patterns reflect the relative position
of each entry not only in the full image, but also to the ROI.

The goal of M-FRCN-g is to exploit the global context -- combining
all the visual patterns in the full image to handle the recognition
of an ROI. Certainly, the visual patterns inside and outside an ROI
may contribute differently to the recognition of an ROI. We expect
that the new masks learned by the proposed MWN-g can reflect the contributions
of these visual patterns. In the inference stage, the learned mask
is independent with the input image and is only relevant to the relative
position of a considered ROI in the image.

\subsection{M-FRCN-lg: Combining M-FRCN-l and M-FRCN-g}

M-FRCN-l and M-FRCN-g extract ROI features in different scales. M-FRCN-l
focuses on ROI's local appearance, while M-FRCN-g exploits global
context. We can combine them to further boost the object-detection
accuracy. In the combined model, the $D_{conv}$-d feature vectors
after GMP from M-FRCN-l and M-FRCN-g are simply concatenated together
and fed to the following layers. Besides, backbone convolutional layers
and RPN are shared by M-FRCN-l and M-FRCN-g. We refer to this combined
network as M-FRCN-lg.

\section{Experiments}

\subsection{Parameters}

\begin{singlespace}
The parameters that need to be tuned in our approach are:

- $D_{fc}$: the number of hidden nodes of the FC layer.

- $I^{l}$, $I_{in}^{g}$, and $I_{out}^{g}$: entry values of the
raw mask.

- $N'$: initial ROI pooling scale for M-FRCN and kernel size for
MWN convolution.

In our experiments, we identify the value of $N'$ at the end of a
model's name, e.g., M-FRCN-l-$7$ indicating the M-FRCN-l with $N'=7$.
For M-FRCN-lg, we always combine M-FRCN-l and M-FRCN-g with the same
$N'$ value, e.g., M-FRCN-lg-$7$ is the combination of M-FRCN-l-$7$
and M-FRCN-g-$7$. All proposed and comparison methods are implemented
using Caffe \cite{Jia2014} and we use a single Titan-X GPU for both
training and inference. All experimental results of the existing works
are re-produced in the same training and inference configurations,
and thus they may be not exactly same as reported in papers.
\end{singlespace}

\subsection{Experiments on PASCAL VOC \label{sub:Experiments-on-VOC}}

\textbf{Implementation details}. Following many existing works on
object detection, we evaluate our approach on the PASCAL VOC detection
benchmarks \cite{Everingham2007}, which contain objects from 20 categories.
We train models on the union set of VOC 2007 trainval and VOC 2012
trainval (VOC 07+12 trainval), and test them on VOC 2007 test set
(VOC 07 test). All models are finetuned on pre-trained ImageNet classification
models. The input image is scaled such that its shorter side is $600$
pixels. We use a weight decay of $0.0005$, a momentum of $0.9$,
and a mini-batch size of $1$. The learning rate is initialized as
$0.001$ and is decreased by a factor of $10$ after $80$k iterations.
A total of $110$k iterations are performed for training. Note that
our MWNs are jointly trained with other network components in an end-to-end
manner. In the inference stage, we still use a single-scale ($600$
pixels) scheme. RPN provides $300$ region candidates for the following
classification and bounding-box regression. A standard non-maximum
suppression (NMS) is performed on the detections with an overlap threshold
of $0.3$. For M-FRCN, we set $I^{l}=1$, $I_{in}^{g}=1$, and $I_{out}^{g}=-1$.
Besides, we set the value of $D_{fc}$ for M-FRCN-l/g/lg according
to the adopted CNN backbone architecture. We report Average Precision
using IoU threshold at $0.5$ (AP@$0.5$) to evaluate the accuracy. 

\textbf{Baselines.} In order to clearly show the effectiveness of
our MWNs and M-FRCNs, we introduce two baseline networks. One  has
a same architecture as M-FRCN-l, except that the original MWN-l is
bypassed and the raw unary mask is directly applied to the ROI feature
map. This baseline network, referred to as baseline-local, is actually
a standard Faster R-CNN with $1\times1$ ROI pooling. The other baseline
has a same architecture as M-FRCN-lg, except that the full-image feature
map is globally pooled directly without using the masking operation
based on MWN-g. This baseline network, referred to as baseline-global,
aims to validate the effectiveness of our MWN-g designed for capturing
global context for each ROI.

\begin{table}
\caption{Accuracy of baselines and M-FRCN-l/g/lg on VOC 07 test, using different
values of $N'$. VGG-16 is used as CNN backbone. \label{tab:different N' and baseline}}

\centering{}{\small{}}%
\begin{tabular}{c|c|c|c}
\hline 
\multicolumn{4}{c}{{\small{}AP@$0.5$ (\%)}}\tabularnewline
\hline 
\hline 
{\small{}baseline-local} & \multicolumn{3}{c}{{\small{}65.0}}\tabularnewline
\hline 
{\small{}baseline-global} & {\small{}74.0} & {\small{}74.7} & {\small{}74.4}\tabularnewline
\hline 
{\small{}M-FRCN-l} & {\small{}73.0} & {\small{}73.7} & {\small{}73.5}\tabularnewline
{\small{}M-FRCN-g} & {\small{}55.6} & {\small{}66.9} & {\small{}68.5}\tabularnewline
{\small{}M-FRCN-lg} & \textbf{\small{}75.4} & \textbf{\small{}75.9} & \textbf{\small{}75.8}\tabularnewline
\hline 
{\small{}$N'$} & {\small{}$7$} & {\small{}$15$} & {\small{}$21$}\tabularnewline
\hline 
\end{tabular}
\end{table}

\textbf{Comparisons with baselines using different $N'$ values.}
At first, we compare M-FRCN-l/g/lg and baselines, by varying the values
of $N'$. In this experiment, VGG-16 \cite{Simonyan2014} is used
as CNN backbone. We set $D_{fc}=256$ for M-FRCN-l and M-FRCN-g, and
$D_{fc}=512$ for M-FRCN-lg. From Tab. \ref{tab:different N' and baseline},
we can see that M-FRCN-l-7/15/21 have much higher AP than baseline-local,
thanks to the proposed channel-wise masking. Besides, M-FRCN-l-7/15/21
have similar APs, suggesting that finer initial ROI-pooling does not
help. Although M-FRCN-g has much lower AP than M-FRCN-l, it is still
interesting to see that M-FRCN-g alone is able to detect objects effectively
using the masked full-image feature map. We conjecture the lower AP
is caused by full-image initial pooling that only preserves very coarse
spatial information. Moreover, M-FRCN-g-15 achieves $11.3\%$ higher
AP than M-FRCN-g-7, and M-FRCN-g-21 performs $1.6\%$ better than
M-FRCN-g-15. This indicates that a fine full-image feature map is
important for M-FRCN-g. Finally, M-FRCN-lg is evaluated. We find that
M-FRCN-lg always outperforms M-FRCN-l and baseline-global, demonstrating
that global context is effectively utilized to improve object-detection
accuracy. Particularly, the comparison between M-FRCN-lg and baseline-global
clearly validates that our ROI-specific context feature leads to better
accuracy than using uniform global feature for all ROIs in an image. 

\textbf{Comparisons with Faster R-CNN and R-FCN using relatively shallow
networks.} We compare our M-FRCN-lg with the standard Faster R-CNN
\cite{Ren2015} using VGG-16 \cite{Simonyan2014} and VGG\_CNN\_M\_1024
\cite{Chatfield2014} models, and R-FCN using VGG16 model. We set
$D_{fc}=256$ for M-FRCN-l and $D_{fc}=512$ for M-FRCN-lg, for both
VGG-16 and VGG\_CNN\_M\_1024. The results are shown in Tab. \ref{tab:comparisons with faster r-cnn}.
We can see that M-FRCN-l-15 has lower AP than standard Faster R-CNN,
but the model complexity is significantly reduced without using network
compression technique and specifically designed CNN backbone. Moreover,
M-FRCN-lg-15 outperforms M-FRCN-l-15 by considering global context
and achieves quite similar accuracy as the standard Faster R-CNN,
with small increase in the number of model parameters. Finally, fully-convolutional
R-FCN has much lower AP than Faster R-CNN and M-FRCNs.

\begin{table}
\caption{Accuracy, model complexity, and test time on VOC 07 test set, using
VGG\_CNN\_M\_1024 and VGG-16 as CNN backbone. \label{tab:comparisons with faster r-cnn}}

\begin{centering}
{\small{}}%
\begin{tabular}{c|c|c}
\hline 
 & {\small{}AP@$0.5$ (\%)} & {\small{}\# params}\tabularnewline
\hline 
\multicolumn{3}{c}{{\small{}VGG\_CNN\_M\_1024 \cite{Chatfield2014}}}\tabularnewline
\hline 
{\small{}Faster R-CNN \cite{Ren2015}} & {\small{}64.8} & {\small{}87.5M}\tabularnewline
{\small{}M-FRCN-l-15} & {\small{}62.1} & {\small{}8.0M}\tabularnewline
{\small{}M-FRCN-lg-15} & {\small{}63.6} & {\small{}8.5M}\tabularnewline
\hline 
\multicolumn{3}{c}{{\small{}VGG-16 \cite{Simonyan2014}}}\tabularnewline
\hline 
{\small{}Faster R-CNN \cite{Ren2015}} & {\small{}76.5} & {\small{}137.1M}\tabularnewline
{\small{}R-FCN \cite{Dai2016}} & {\small{}62.3} & {\small{}17.6M}\tabularnewline
\hline 
{\small{}M-FRCN-l-15} & {\small{}73.7} & {\small{}17.4M}\tabularnewline
{\small{}M-FRCN-lg-15} & {\small{}75.9} & {\small{}17.9M}\tabularnewline
\hline 
\end{tabular}
\par\end{centering}{\small \par}

\centering{}
\vspace{-0.00in}
\end{table}

\begin{table*}
\caption{Accuracy, model complexity, and test time of R-FCNs and M-FRCNs on
COCO test-dev set, using ResNet-101 or SE-ResNet-101 as CNN backbone.
\label{tab:Comparison with R-FCN} }

\centering{}{\small{}}%
\begin{tabular}{l|l|c|c|ccc|c|c}
\hline 
{\small{}CNN backbone} & {\small{}feature encoder } & {\small{}AP@$0.5$} & {\small{}AP} & {\small{}AP} & {\small{}AP} & {\small{}AP} & {\small{}\# params} & {\small{}run-time}\tabularnewline
 &  &  &  & {\small{}small} & {\small{}medium} & {\small{}large} &  & {\small{}(sec/img)}\tabularnewline
\hline 
{\small{}ResNet-101} & {\small{}PS ROI pooling \cite{Dai2016}} & {\small{}48.9} & {\small{}28.8} & {\small{}10.7} & {\small{}31.2} & {\small{}41.8} & {\small{}53.8M} & {\small{}0.15}\tabularnewline
{\small{}\cite{He2016}} & {\small{}deformable PS ROI pooling \cite{Dai2017}} & {\small{}50.0} & {\small{}30.6} & {\small{}11.9} & {\small{}34.1} & {\small{}43.7} & {\small{}62.1M} & {\small{}0.23}\tabularnewline
\cline{2-9} 
 & {\small{}our M-FRCN-l-7} & {\small{}51.0} & {\small{}30.7} & {\small{}11.6} & {\small{}33.4} & {\small{}45.0} & {\small{}49.5M} & {\small{}0.15}\tabularnewline
 & {\small{}our M-FRCN-lg-7} & {\small{}51.9} & {\small{}31.2} & {\small{}12.0} & {\small{}34.3} & {\small{}46.3} & {\small{}51.7M} & {\small{}0.17}\tabularnewline
\hline 
\hline 
{\small{}SE-ResNet-101} & {\small{}PS ROI pooling \cite{Dai2016}} & {\small{}51.3} & {\small{}30.7} & {\small{}11.6} & {\small{}33.7} & {\small{}44.8} & {\small{}58.6M} & {\small{}0.16}\tabularnewline
{\small{}\cite{Hu2017}} & {\small{}deformable PS ROI pooling \cite{Dai2017}} & {\small{}52.8} & {\small{}32.6} & {\small{}12.8} & {\small{}36.3} & {\small{}46.7} & {\small{}66.8M} & {\small{}0.24}\tabularnewline
\cline{2-9} 
 & {\small{}our M-FRCN-l-7} & {\small{}53.6} & {\small{}33.0} & {\small{}12.5} & {\small{}36.4} & {\small{}48.3} & {\small{}54.3M} & {\small{}0.16}\tabularnewline
 & {\small{}our M-FRCN-lg-7} & {\small{}54.3} & {\small{}33.4} & {\small{}12.8} & {\small{}36.6} & {\small{}49.0} & {\small{}56.5M} & {\small{}0.18}\tabularnewline
\hline 
\end{tabular}
\end{table*}

\subsection{Experiments on MS COCO}

\textbf{Implementation details}. We also evaluate our M-FRCNs on the
MS COCO dataset \cite{Lin2014a}, that contains images of 80 categories
of objects. We train models using the union of training and validation
images (COCO trainval), and test on the test-dev set (COCO test-dev)
\cite{Lin2016}. In this experiment, we use the similar training and
inference configurations as the experiment on Pascal VOC, except that
the learning rate is initialized as $0.0005$ and is decreased by
a factor of $10$ after $1.28$M iterations. Totally, $1.92$M iterations
are performed for training. Besides, the online hard example mining
(OHEM) \cite{Shrivastava2016} is also adopted for training. We evaluate
object detection by COCO-style AP @ $IoU\in[0.5,0.95]$ and PASCAL-style
AP@ $IoU=0.5$. Model complexity and run-time efficiency are also
reported.

\textbf{Comparisons with R-FCNs using very deep networks.} We conducted
experiments by using very deep networks to build our M-FRCNs and compare
their performance against two state-of-art feature encoders for object-detection
networks. One is Position-Sensitive (PS) ROI pooling proposed in R-FCN
\cite{Dai2016}, and the other one is deformable PS ROI pooling which
is proposed in DCN \cite{Dai2017}. Deformable PS ROI pooling allows
grids to shift from their original positions and thus becomes much
more flexible, but the bilinear interpolation operations performed
on score maps lead to slower run-time, especially when the number
of categories is large. As for CNN backbone, we choose to use ResNet-101
\cite{He2016} and SE-ResNet-101 \cite{Hu2017}. SE-ResNet-101 equips
ResNet-101 with Squeeze-and-Excitation components, which is the key
technique of the winner model of ILSVRC 2017 Image Classification
Challenge. Moreover, we set $D_{fc}=1024$ for M-FRCN-l and M-FRCN-g,
and $D_{fc}=2048$ for M-FRCN-lg. $N'$ is set to $7$ to fit model
training to $12$GB GPU memory. Note that DCN is officially implemented
in a different deep-learning framework (MXNet) with ours (Caffe),
so we re-implement the deformable PS ROI pooling in Caffe. But we
did not re-implement the deformable CNN backbone, since this work
only focuses on feature encoder. We expect our current experiments
can show the ability of our proposed feature encoder to improve object-detection
performance for different CNN backbones.

The results are shown in Tab. \ref{tab:Comparison with R-FCN}. Compared
with the standard R-FCN using PS ROI pooling, M-FRCN-ls have higher
accuracy, lower model complexity and similar run-time efficiency.
Moreover, M-FRCN-ls have a little higher overall AP and about $30\%$
faster run-time efficiency than the enhanced R-FCN using deformable
PS ROI pooling, with fewer parameters. M-FRCN-lg-7 further improves
the accuracy of M-FRCN-l-7 by considering global context, while the
number of parameters increases by $2.2$M and run-time increases by
$0.02$ sec/img. Compared with the enhanced R-FCN, M-FRCN-lg-7 has
higher object-detection accuracy, less run-time, and lower model complexity. 

\begin{figure}
\begin{centering}
\includegraphics[scale=0.4]{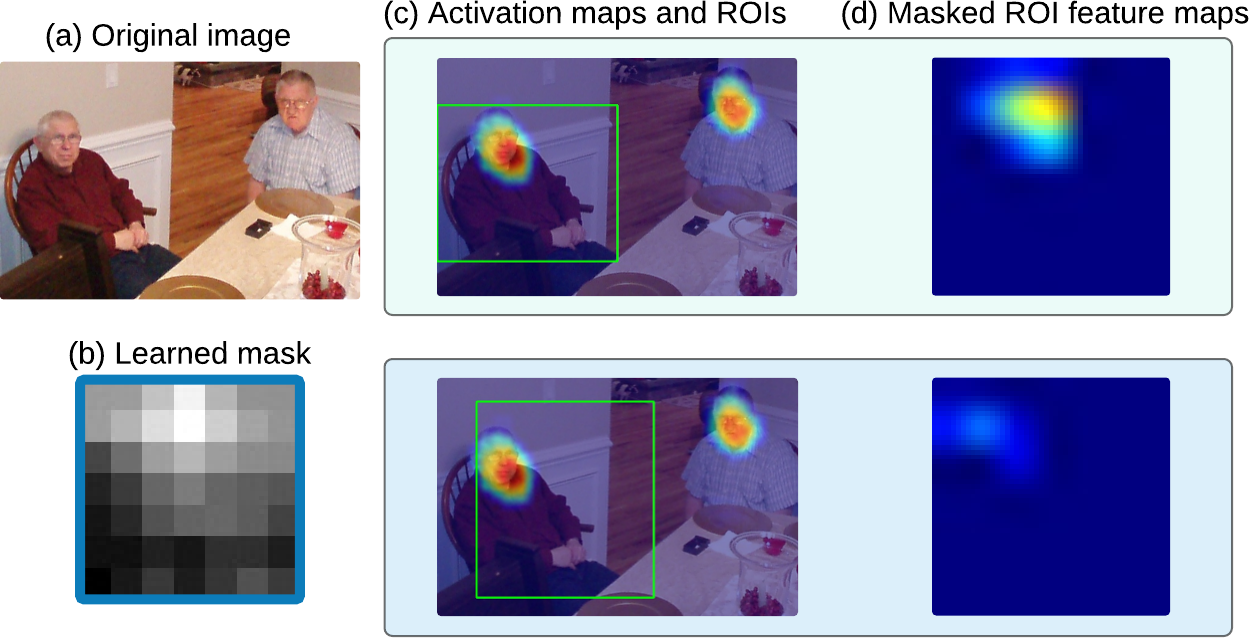}
\par\end{centering}

\caption{An example of how M-FRCN-l encodes spatial information for a feature
channel which is strongly activated for human head. (a) Original image.
(b) Mask learned by MWN-l for this channel. (c) Visualization of the
channel activation and two example ROIs (indicated by green boxes).
(d) Resulting ROI feature maps after masking. \label{fig:MWN-l_head_mask}}
\end{figure}

\begin{figure}
\begin{centering}
\includegraphics[scale=0.51]{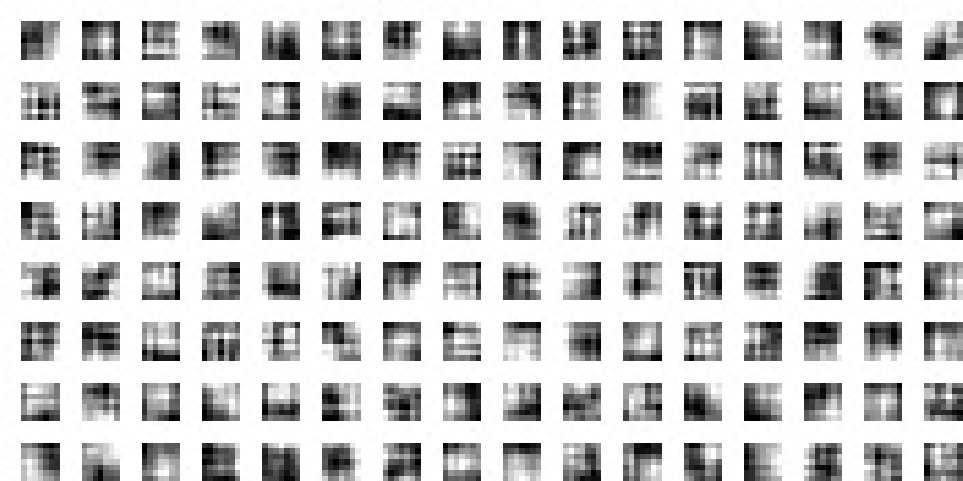}
\par\end{centering}

\caption{128 example masks learned by MWN-l. \label{fig:128-example-masks}}
\end{figure}

\subsection{Qualitative Analysis}

In this section, we use real examples to show how the masks learned
by MWN-l and MWN-g help object detection. In Fig. \ref{fig:MWN-l_head_mask},
we present an example of how M-FRCN-l encodes spatial information
for a feature channel which is strongly activated for human head.
The mask learned by MWN-l for this channel has high values at the
top-center positions as shown in Fig. \ref{fig:MWN-l_head_mask}(b),
which is reasonable since head tends to appear at the top-center position
of a human body in COCO dataset. By multiplying the `head' channel
of a ROI feature map (Fig.\ref{fig:MWN-l_head_mask}(c)) with its
mask, the spatial information of human head can be effectively encoded
for object detection. As shown in Fig. \ref{fig:MWN-l_head_mask}(c)
and (d), if an ROI accurately overlaps with a human body, there is
strong activation in the masked feature. On the contrary, if an ROI
does not overlap with a human body very well, the activation is very
weak. Moreover, we present Fig. \ref{fig:128-example-masks} to show
128 example masks learned by MWN-l. We can see that our MWN-l can
produce masks with very complicated patterns. 

\begin{figure}
\begin{centering}
\includegraphics[scale=0.35]{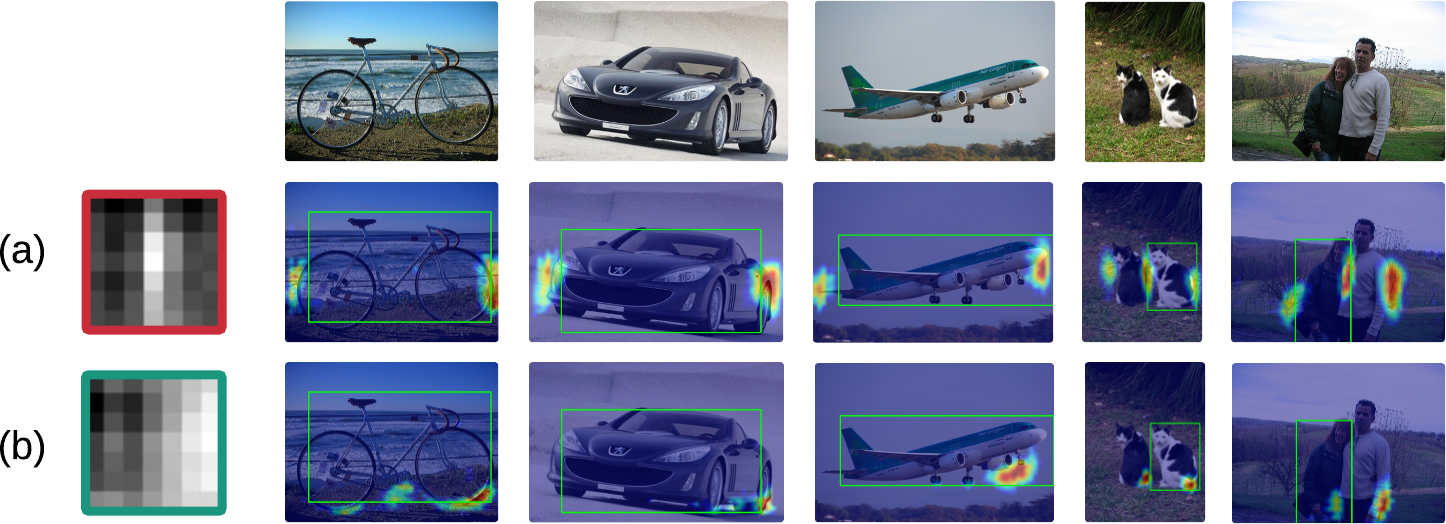}
\par\end{centering}

\caption{Rows (a) and (b) show the activation maps of two channels on five
images, and the masks learned for these channels. Original images
are shown on the top row and example ROIs are indicated by green boxes.
\label{fig:localization channels}}
\end{figure}

For M-FRCN-l, we also observe feature channels that are activated
for specific position of an object, rather than specific visual pattern.
We think these feature channels can help object localization. Fig.
\ref{fig:localization channels}(a) and (b) present activation maps
of two such channels and their learned masks. The feature channel
shown in Fig. \ref{fig:localization channels}(a) is strongly activated
for the horizontal endpoints of an object. Interestingly, the learned
mask for this feature channel has high values in the middle, rather
than the left and the right sides. We believe the use of this mask
can help detect an object by not confusing with other nearby objects,
as shown in the fourth and fifth example images in Fig. \ref{fig:localization channels}(a).
The feature channel shown in Fig. \ref{fig:localization channels}(b)
is strongly activated for the lower right corner of an object, and
the learned mask has high values at the right side, which is also
reasonable. 

\begin{figure}
\begin{centering}
\includegraphics[scale=0.47]{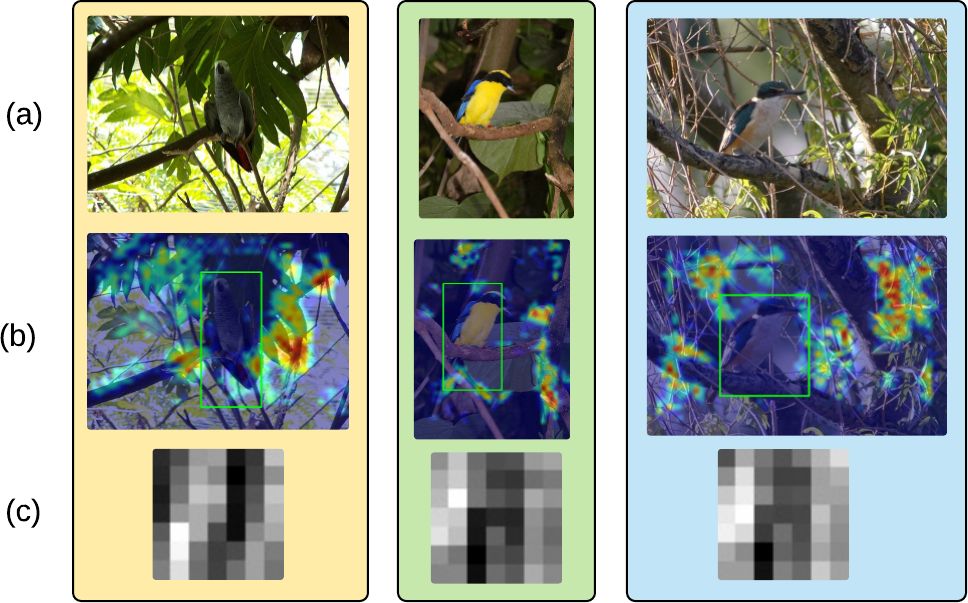}
\par\end{centering}

\caption{Examples of masks learned in M-FRCN-g for a feature channel which
is strongly activated at leaves around a bird. (a) Original images.
(b) Visualization of the channel activation and example ROIs (indicated
by green boxes). (c) Masks produced by MWN-g for this channel, given
the ROIs presented in (b).  \label{fig:MWN-g_bird_mask}}
\end{figure}

For M-FRCN-g, we present three examples in Fig. \ref{fig:MWN-g_bird_mask}
to show the masks learned by MWN-g for a feature channel which is
strongly activated at leaves around a bird. Fig. \ref{fig:MWN-g_bird_mask}(a)
shows three bird images where birds are surrounded by leaves, and
Fig. \ref{fig:MWN-g_bird_mask}(b) presents the activation maps for
the `leaf' feature channel and ROI examples. As shown in Fig. \ref{fig:MWN-g_bird_mask}(c),
the learned masks largely have high values around the ROI, indicating
that \textbf{the leaves} (visual pattern) \textbf{around a ROI} (spatial
information) can contribute to the ROI recognition. We can see that
our M-FRCN-g provides a novel method to utilize the visual patterns
outside the ROI in the full image to serve for the recognition
of a specific ROI.

\begin{figure*}
\begin{centering}
\includegraphics[scale=0.4]{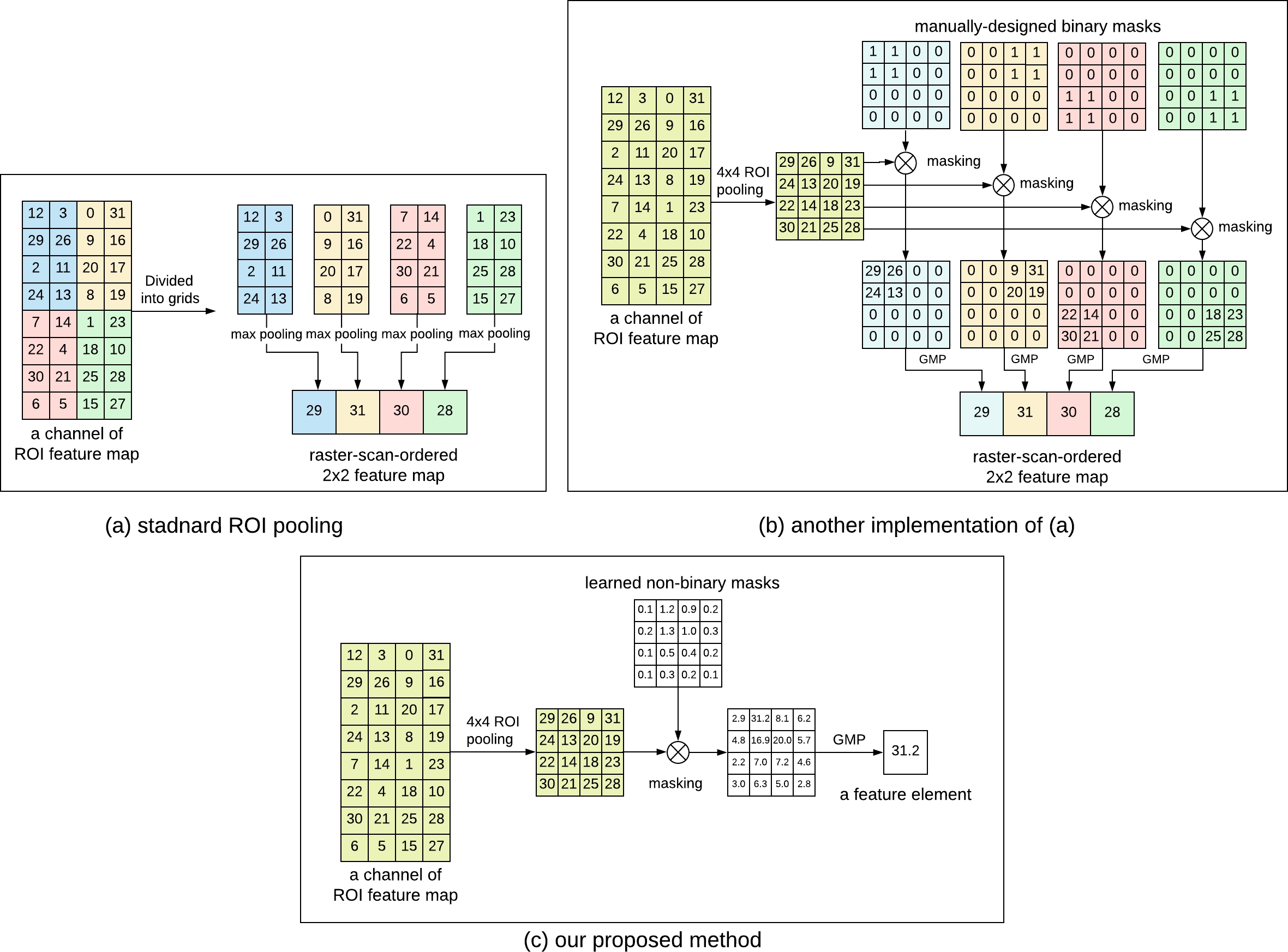}
\par\end{centering}

\caption{An illustration of ROI pooling by including masking operations. (a)
ROI pooling ($2\times2$) in existing R-CNNs, where grids are represented
by different colors. (b) Using a set of masks to implement the $2\times2$
ROI pooling shown in (a), by sequentially performing an initial $4\times4$
ROI pooling, four masking operations, and global max poolings (GMPs).
(c) In this paper, our basic idea is to relax these masks to non-binary
ones and learn them automatically for better feature encoding and
object detection. \label{fig:Traiditional-RoI-pooling}}
\end{figure*}

\section{Relation between Masked-based and Grid-based Feature Encoders}

In this section, we show that the proposed mask-based feature encoder
is a generalization of the traditional grid-based one. In the standard
Fast/Faster R-CNN architectures, ROI pooling partitions each region
proposal into $N\times N$ equal-sized grids and then performs a pooling
operation within each grid to produce a fixed-dimensional ROI feature
map for the following layers. The ROI feature map is transformed to
a feature vector in the raster-scan order, which reflects the spatial
relations between grids. An example of this ROI pooling operation
with $N=2$ is shown in Fig. \ref{fig:Traiditional-RoI-pooling}(a).
Actually, this standard ROI pooling can be implemented by an initial
$N'\times N'$ pooling with $N'>N$ (e.g. $N'=4$) followed by applying
a set of $N'\times N'$ \textbf{binary} masks and global max poolings,
as shown in Fig. \ref{fig:Traiditional-RoI-pooling}(b). In each mask,
only entries in specific spatial locations take the value 1. 

Our mask-based feature encoder relaxes these masks to more informative
\textbf{non-binary} ones and learn them using a MWN. Furthermore,
it can be relaxed to learn different sets of masks for different channels.
Fig. \ref{fig:Traiditional-RoI-pooling}(c) shows an example of masking
operation in the proposed method, where a learned non-binary mask
is applied to a channel of ROI feature map, and a feature element
is output.

\section{Conclusion}

In this paper, we proposed a new feature encoder for object detection.
Unlike the existing methods which utilize grids to represent object-parts
and learn what is likely to appear in each grid, the proposed method
learns masks to reflect spatial distributions of a set of visual patterns.
The proposed feature encoder can be used to capture both local ROI
appearance and global context. By integrating our feature encoder
to Faster R-CNN architecture, we obtain MWN-based Faster R-CNN (M-FRCN).
As shown by the experimental results, M-FRCNs have comparable object-detection
accuracy with the standard Faster R-CNNs, but have significantly reduced
model complexity. When compared with R-FCNs using very-deep CNN backbones,
M-FRCNs can produce higher object-detection accuracy with good run-time
efficiency.

\end{document}